\documentclass[runningheads]{llncs}
\usepackage[T1]{fontenc}

\usepackage{graphicx} %
\usepackage{verbatim}
\usepackage{pifont}
\usepackage{ifthen}

\usepackage[table]{xcolor}
\usepackage{amsmath}
\usepackage{amsfonts}
\usepackage{subcaption}

\usepackage{booktabs}   %
\usepackage{multicol}   %
\usepackage{multirow}   %
\usepackage{makecell}   %
\usepackage{caption}

\usepackage{hyperref}
\usepackage{cleveref}

\usepackage{graphicx,verbatim}
\renewcommand{\paragraph}[1]{{\vspace{1mm}\noindent \bf #1}.}

\begin{document}
\title{Mitigating Biases in Surgical Operating Rooms with Geometry}

\author{
Tony Danjun Wang \inst{1} \and 
Tobias Czempiel \inst{3} \and 
Christian Heiliger \inst{4} \and \\
Nassir Navab \inst{1,2} \and
Lennart Bastian \inst{1,2}
}  
\authorrunning{T. Wang et al.}
\institute{Computer Aided Medical Procedures, TU Munich, Germany \and
Munich Center for Machine Learning, Germany \and
UCL Hawkes Institute, Dept. Computer Science, University College London \and
Minimally Invasive Surgery, University Hospital of Munich (LMU), Germany
}

\maketitle              %
\begin{abstract}
\vspace{-.14cm}
Deep neural networks are prone to learning spurious correlations, exploiting dataset-specific artifacts rather than meaningful features for prediction.
In surgical operating rooms (OR), these manifest through the standardization of smocks and gowns that obscure robust identifying landmarks, introducing model bias for tasks related to modeling OR personnel.
Through gradient-based saliency analysis on two public OR datasets, we reveal that CNN models succumb to such shortcuts, fixating on incidental visual cues such as footwear beneath surgical gowns, distinctive eyewear, or other role-specific identifiers.
Avoiding such biases is essential for the next generation of intelligent assistance systems in the OR, which should accurately recognize personalized workflow traits, such as surgical skill level or coordination with other staff members.
We address this problem by encoding personnel as 3D point cloud sequences, disentangling identity-relevant shape and motion patterns from appearance-based confounders.
Our experiments demonstrate that while RGB and geometric methods achieve comparable performance on datasets with apparent simulation artifacts, RGB models suffer a 12\% accuracy drop in realistic clinical settings with decreased visual diversity due to standardizations.
This performance gap confirms that geometric representations capture more meaningful biometric features, providing an avenue to developing robust methods of modeling humans in the OR.


\end{abstract}
\section{Introduction}
\label{sec:introduction}

\vspace{-1mm}

Deep neural networks excel at abstracting statistical patterns in data, but this capability becomes problematic when models learn spurious correlations instead of meaningful features \cite{geirhosImageNettrainedCNNsAre2022}.
Data-driven models exploit dataset-specific artifacts known as shortcuts \cite{geirhos2020shortcut}; this lack of robustness introduces vulnerabilities that must be avoided in safety-critical domains such as surgical operating rooms (OR).
In clinical environments, these issues are particularly prominent due to limited appearance variations \textit{within} but large variations \textit{between} clinics.

\begin{figure}[t]
    \centering
    \includegraphics[width=\textwidth]{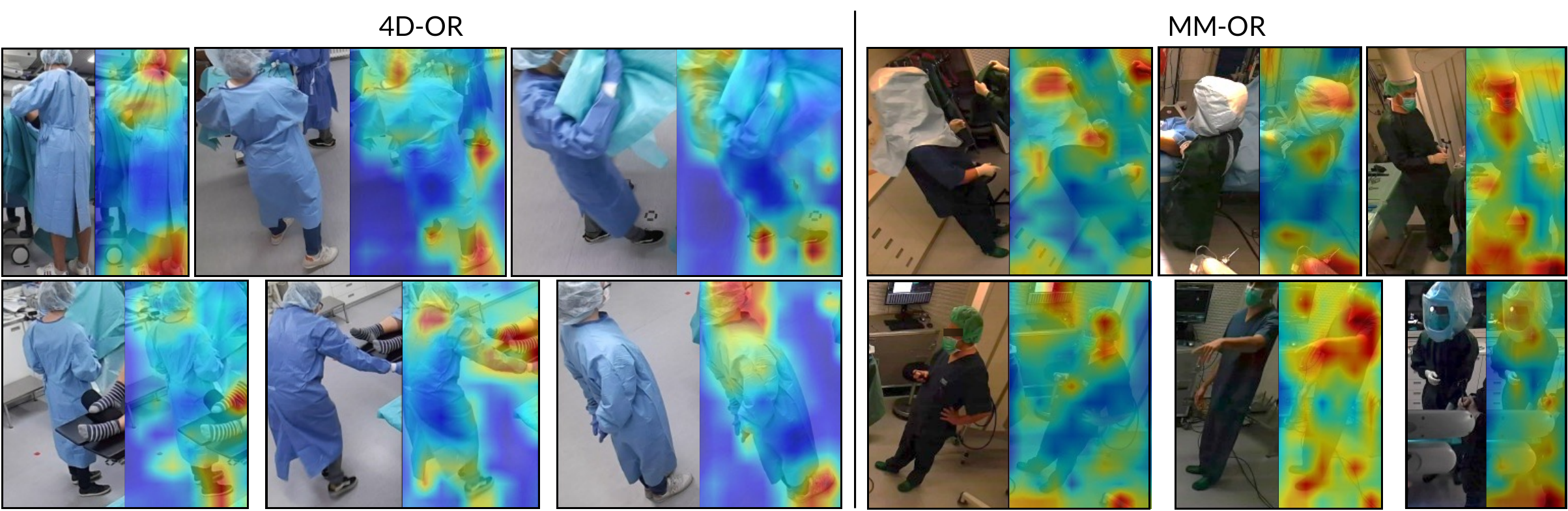}
    \caption{RGB image excerpts and overlayed saliency maps generated with GradCAM \cite{selvaraju2017grad} on the simulated datasets 4D-OR \cite{ozsoy20224d} and MM-OR \cite{ozsoy2025mmor}.
    4D-OR's limited realism and variety allow CNNs to identify individuals solely by their heads and shoes. In more realistic OR settings like MM-OR, these features become less useful due to more homogenous attire. These discrepancies between different clinical environments can impede generalization.}
    \label{fig:saliency_maps}
    \vspace{-6.0mm}
\end{figure}

Standardized attire, including scrubs, masks, and protective equipment, removes the diversity of visual features that computer vision models typically exploit for the modeling of humans in the OR \cite{liInDepthExplorationPerson2023,srivastav2018mvor,bastian2023disguisor,liu2024human,bastian2023know}.
When distinctive features are obscured, models resort to learning spurious correlations with whatever visual artifacts remain available (see \cref{fig:saliency_maps}).

Through gradient-based saliency analysis \cite{selvaraju2017grad}, we uncover how CNNs trained to \textbf{re-identify OR personnel} exhibit systematic biases toward incidental visual cues.
Rather than learning robust biometric features, these models fixate on simulation artifacts in 4D-OR \cite{ozsoy20224d} such as street shoes visible beneath surgical gowns, or distinct eyewear.
Such features may correlate with identity within a specific dataset but represent spurious shortcuts that fail when deployed in new clinical environments.

Our analysis suggests that bias mitigation in constrained visual environments stands to benefit from a shift in representation learning.
Rather than attempting to debias RGB features, we advocate for geometric modeling through 3D point cloud sequences.
For the task of person re-identification, RGB features are not sufficiently useful due to the indiscriminate nature of standardized surgical attire.
On the other hand, point cloud sequences naturally emphasize invariant geometric properties such as shape, gait dynamics, and movement patterns that persist across appearance changes \cite{fanOpenGaitRevisitingGait2023,lidargait,wang2025beyond}.
In this extended abstract, we briefly highlight the relevant contributions from our full Medical Image Analysis Publication \textit{Beyond Role-based Surgical Domain Modeling} \cite{wang2025beyond}.

\section{Methods}
\label{sec:methods}

\vspace{-1mm}

\paragraph{Saliency Analysis of RGB Features} We trained ResNet-50 models for person re-identification on two surgical datasets: the simulated 4D-OR \cite{ozsoy20224d} and the more realistic MM-OR \cite{ozsoy2025mmor}. Using GradCAM \cite{selvaraju2017grad}, we generated class activation maps to visualize which spatial regions the networks relied upon for identity discrimination. 
Models are trained on a supervised classification task of the individuals' known identity.

\begin{figure}[t]
    \centering
    \includegraphics[width=\columnwidth]{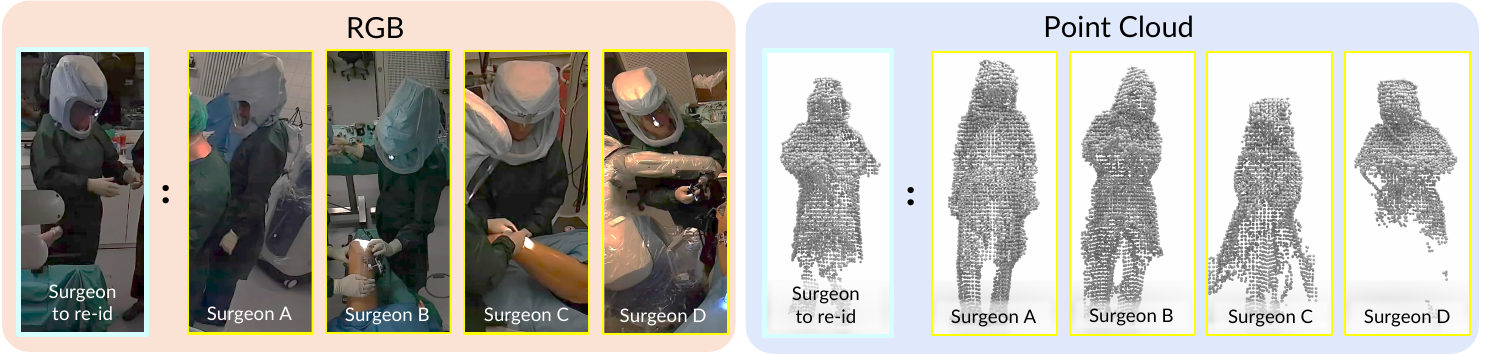}
    \caption{
    \textit{Who is the surgeon to re-identify?}
    Four surgeons (persons A, B, C, and D) are challenging to visually differentiate in RGB images.
    They can be more easily distinguished based on their height differences, which is ambiguous in RGB images due to the loss of absolute scale.
    \textit{The correct answer is Person D.}
    }
    \label{fig:teaser}
    \vspace{-7.0mm}
\end{figure}

\paragraph{Color vs. Geometry Ablation}
To quantify the contribution of appearance versus shape information, we conduct ablations comparing RGB and depth representations within the LiDARGait framework \cite{lidargait}. 
Individuals are first segmented from the 3D scene using a weakly-supervised approach \cite{bastian2023segmentor}.
Then, we modify LiDARGait to ingest RGB in a manner identical to the rendered depth sequences.
Models are trained using standard triplet loss with online hard negative mining \cite{leibe_triplet}.
All experiments compare RGB and point cloud-based representations within identical network architectures to isolate the contribution of input modality.

\paragraph{Datasets and Experimental Design}
We evaluate our approach on two surgical datasets: the simulated 4D-OR \cite{ozsoy20224d} containing 5 individuals across 10 procedures, and the more authentic MM-OR \cite{ozsoy2025mmor} containing 13 individuals across 11 robotic knee surgeries.
While MM-OR does not have real patients, it is more authentic as real clinicians perform the surgeries.
We conduct four-fold cross-validation, partitioning by individual surgeries to ensure temporal separation between training and testing data.
Following standard re-identification protocols \cite{reid_review}, we adopt a probe-gallery evaluation framework where gallery sequences of known identities are compared against probe sequences of unknown identities.

\paragraph{Metrics and Statistical Analysis}
Performance is assessed using established re-identification metrics: mean Average Precision (mAP), rank-3 cumulative matching characteristic (CMC@3), and rank-1 micro and macro accuracy \cite{ming2022deep}.
Statistical significance is determined through repeated-measures ANOVA followed by paired t-tests with Bonferroni correction.

\section{Findings}

\vspace{-1mm}

\paragraph{Saliency Analysis Reveals Dataset-Specific Biases}
GradCAM analysis reveals fundamentally different saliency heatmaps between datasets (see \cref{fig:saliency_maps}).
On 4D-OR, CNNs consistently focus on the head and feet regions across all individuals, exploiting visible streetwear and distinctive footwear beneath surgical gowns.
These artifacts provide strong identity signals within the dataset but represent spurious correlations that would fail in authentic clinical settings.
In contrast, MM-OR saliency maps reveal more diffuse, unfocused patterns lacking clear regions of attention.
This dispersed activation suggests models struggle to identify reliable visual features when more distinguishing characteristics are not present due to standardized surgical attire.

\paragraph{Geometric Representations Outperform Appearance-Based Features}
Quantitative evaluation confirms the limitations of RGB-based identification in realistic OR settings (see \cref{tab:intra_dataset}).
While RGB and point cloud methods achieve comparable performance on 4D-OR (96.91\% vs 95.95\% macro accuracy, $p > 0.05$), their effectiveness diverges dramatically on MM-OR.
Point cloud representations maintain high accuracy (85.75\%) while RGB performance drops significantly to 73.23\% ($p < 0.05$).
This 12\% performance gap demonstrates that geometric features provide more reliable identity signals than appearance when in more realistic clinical settings without obvious simulation artefacts.
The near-perfect RGB performance on 4D-OR further supports our hypothesis that models exploit dataset-specific visual artifacts rather than learning robust biometric features, suggesting the re-identification task is trivially solved.
Cross-dataset evaluation amplifies these differences: models trained on one OR environment and tested on another show catastrophic RGB performance degradation (46.98\% accuracy) while point cloud methods maintain reasonable transfer performance (65.92\%).
These results validate our hypothesis that shifting from appearance to geometric representations mitigates the impact of visual biases in the OR.

\begin{table}[t]
\centering
\caption{Intra-dataset re-identification performance. RGB methods achieve near-perfect accuracy on synthetic 4D-OR but struggle on realistic MM-OR, where depth-based approaches prove more effective. Statistical significance compared to the best-performing method is indicated (*: $p < 0.05$, ns: not significant).}
\label{tab:intra_dataset}
\begin{tabular}{llcccc}
\toprule
\textbf{Method} & \textbf{Dataset} & \textbf{mAP} & \textbf{CMC@3} & \textbf{Acc. (Micro)} & \textbf{Acc. (Macro)} \\
\midrule
Ours (RGB) & 4D-OR & $\mathbf{98.57 \pm 0.30}$ & $\mathbf{98.97 \pm 0.22}$ & $\mathbf{97.24 \pm 0.56}$ & $\mathbf{96.91 \pm 0.72}$ \\
Ours (PC) & 4D-OR & $97.84 \pm 0.19^{\text{ns}}$ & $98.28 \pm 0.18^{\text{*}}$ & $95.87 \pm 0.49^{\text{ns}}$ & $95.95 \pm 0.52^{\text{ns}}$ \\
\midrule
Ours (RGB) & MM-OR & $81.71 \pm 2.57^{\text{*}}$ & $84.02 \pm 2.70^{\text{*}}$ & $69.50 \pm 3.60^{\text{*}}$ & $73.23 \pm 3.61^{\text{*}}$ \\
Ours (PC) & MM-OR & $\mathbf{91.58 \pm 1.54}$ & $\mathbf{92.45 \pm 1.41}$ & $\mathbf{86.19 \pm 2.24}$ & $\mathbf{85.75 \pm 1.95}$ \\
\bottomrule
\end{tabular}
\vspace{-5.0mm}
\end{table}

\section{Concluding Remarks}

Our analysis reveals fundamental challenges for data-driven models in surgical environments: when the diversity of visual features is constrained, models resort to learning spurious correlations with whatever artifacts remain available.
This is particularly evident for humans in the operating room, who dress in strictly standardized ways due to sterility requirements.
We propose to overcome these limitations by learning from unstructured point cloud representations of each individual.
This approach more effectively encapsulates identity through shape and articulated motion patterns, which in the future can be used to develop personalized intelligent systems in the OR \cite{wang2025beyond}.

\clearpage

\bibliographystyle{splncs04}
\bibliography{main}
\end{document}